# Enhanced semi-supervised stamping process monitoring with physically-informed feature extraction

Jianyu Zhang [a]*

[a] *School of Advanced Manufacturing, Sun Yat-sen University, Shenzhen, 518000, China*

**Abstract**

In tackling frequent batch anomalies in high-speed stamping processes, this study introduces a novel semi-supervised in-process anomaly monitoring framework, utilizing accelerometer signals and physics information, to capture the process anomaly effectively. The proposed framework facilitates the construction of a monitoring model with imbalanced sample distribution, which enables in-process condition monitoring in real-time to prevent batch anomalies, which helps to reduce batch defects risk and enhance production yield. Firstly, to effectively capture key features from raw data containing redundant information, a hybrid feature extraction algorithm is proposed to utilize data-driven methods and physical mechanisms simultaneously. Secondly, to address the challenge brought by imbalanced sample distribution, a semi-supervised anomaly detection model is established, which merely employs normal samples to build a golden baseline model, and a novel deviation score is proposed to quantify the anomaly level of each online stamping stroke. The effectiveness of the proposed feature extraction method is validated with various classification algorithms. A real-world in-process dataset from stamping manufacturing workshop is employed to illustrate the superiority of proposed semi-supervised framework with enhance performance for process anomaly monitoring.

*Keywords:* Stamping Process; In-Process Monitoring; Physically-Informed Feature Extraction; Semi-Supervised Learning

## 1. Introduction

Stamping, utilizing presses to shape flat metal sheets, which forms various components such as body panels and engine parts, is recognized as a crucial metalworking process integral to automobile production. Remarkably, more than 40% of over 2,000 automotive components are derived from metal stamping[1], illustrating the central role of stamping in automotive manufacturing. With the rapid advancement in the integration and complexity of stamping processes, the role of process monitoring in stamping has become increasingly crucial for enhancing production efficiency and ensuring the superior quality of automotive products.

Numerous investigations have been performed on monitoring the process or quality of stamping, including traditional manual inspection, optical quality inspection[2], quality enhancement based on Statistical Process Control (SPC)[3], and so on. Specifically, researchers have focused on employing various sensors, such as optical and force sensors, to identify irregularities, cracks, or scratches on the surface of work-pieces[4,5], thus improving the speed and accuracy of detection. The research and application of visual methods have been mainly developed, which are primarily implemented at the end of the line or specific stations, capable of identifying and classifying different types of defects using techniques such as template matching and computer vision [6–9]. Additionally, SPC and some monitoring methods leverage statistical methods and machine learning techniques to analyze process data for patterns and anomalies, enhancing quality by offering detailed insights into the factors that determine quality within the stamping process[10–13]. By accurately identifying and classifying surface defects on work-pieces, these methods ensure the consistency and reliability of products while reducing the rate of waste and repair costs in production.

*E-mail address:* zhangjy685@mail2.sysu.edu.cn.

In the context of advanced and precision manufacturing, existing methods have indeed seen some successful implementations. However, considering of practicality and scalability, there are still several challenges to be addressed:

- Given the increasing amounts of units per order and the enclosed nature of automated process, it is more common to encounter a batch of anomalies rather than a single anomaly, with the batch sometimes containing hundreds or even thousands of pieces, which requires an in-process monitoring feedback.
- Due to the complexity of stamping conditions and external disturbances, particularly noise, and as the complexity of work-piece structures increases, the challenges of monitoring intensify. However, conversely, the precision requirements for monitoring continue to rise. And in the context of increased automation, reducing downtime caused by false alarms is especially prioritized.
- Due to the weak digital infrastructure and limited application of sensing technologies, the necessary data foundation for machine learning and other technologies is quite inadequate. Moreover, considering the stringent quality control requirements in production, the acquisition of anomaly samples is proved to be extremely challenging.
- Existing methods primarily focus on the application of statistical and machine learning techniques, with less considering of physics information. Consequently, it is hard for these approaches to provide a reasonable analysis of the root causes of anomalies, which is crucial for process improvement and the prevention of future problems.

Specifically, quality inspection related methods, which are usually considered as post-process action to sort out defective pieces but incapable to promptly halt production to prevent batch anomalies. These methods do not perform adequately and fail to provide root cause analysis. SPC methods struggle to effectively utilize large amounts of process data and historical knowledge, which limits their ability to achieve better outcomes. As an extension of SPC method with modern mathematic approach, Shi introduced the concept of in-process quality improvement to monitor the whole stamping process[14–16]. However, there are still deficiencies in the research on the physics inform of process signals, particularly in the in-depth analysis study of high-dimensional signals.

To address these issues, considering the use of process big data and physically-informed approaches, this paper proposes a framework for anomaly monitoring in stamping process using accelerometer data. First, for the large amounts of accelerometer data collected from process, a method for data enhanced processing and feature engineering is proposed. This involves the preprocessing of signals using digital signal processing techniques, which include suppressing noise and signal drift. Furthermore, the extraction of key features is based on the integration of data-driven methods and physically-informed approaches. Secondly, to tackle the challenges of large datasets and the imbalance between positive and negative samples, a semi-supervised learning method is proposed. With experience in data handling and key feature processing, a golden baseline model utilizing only normal data is constructed, enable the identification and quantification of anomaly with a specific score. Ultimately, by employing a variety of models from statistical method, Machine Learning (ML), and Deep Learning (DL), the features are selected and the effectiveness of the hybrid feature extraction method is validated.

The rest of this article is organized as follows: Section I highlights the challenges encountered in the manufacturing process. Section II elaborates on data processing and feature engineering, integrating multiple models for feature selection and validation of the work flow. Section III describes the construction of the golden baseline model and explores methods for quantifying anomalies. Finally, Section IV concludes the article.

## 2. Technical Approach

To validate the data processing and feature extraction workflow for stamping anomaly monitoring, the experimental procedure is illustrated in Figure 1. Initially, the stamping acceleration signals are processed using a Butterworth filter to remove high-frequency noise. Subsequently, combining physics inform with data-driven methods, segmental statistical features and principal components from Principal Component Analysis (PCA) are extracted from the filtered signals. Feature set is appropriately optimized using a model-based feature selection approach. Then considering the extreme imbalance of normal and abnormal samples in stamping scenario, the golden baseline model is constructed with normal data. The model provides a metric score, which could quantify the deviation of anomaly and make normal or abnormal decision.

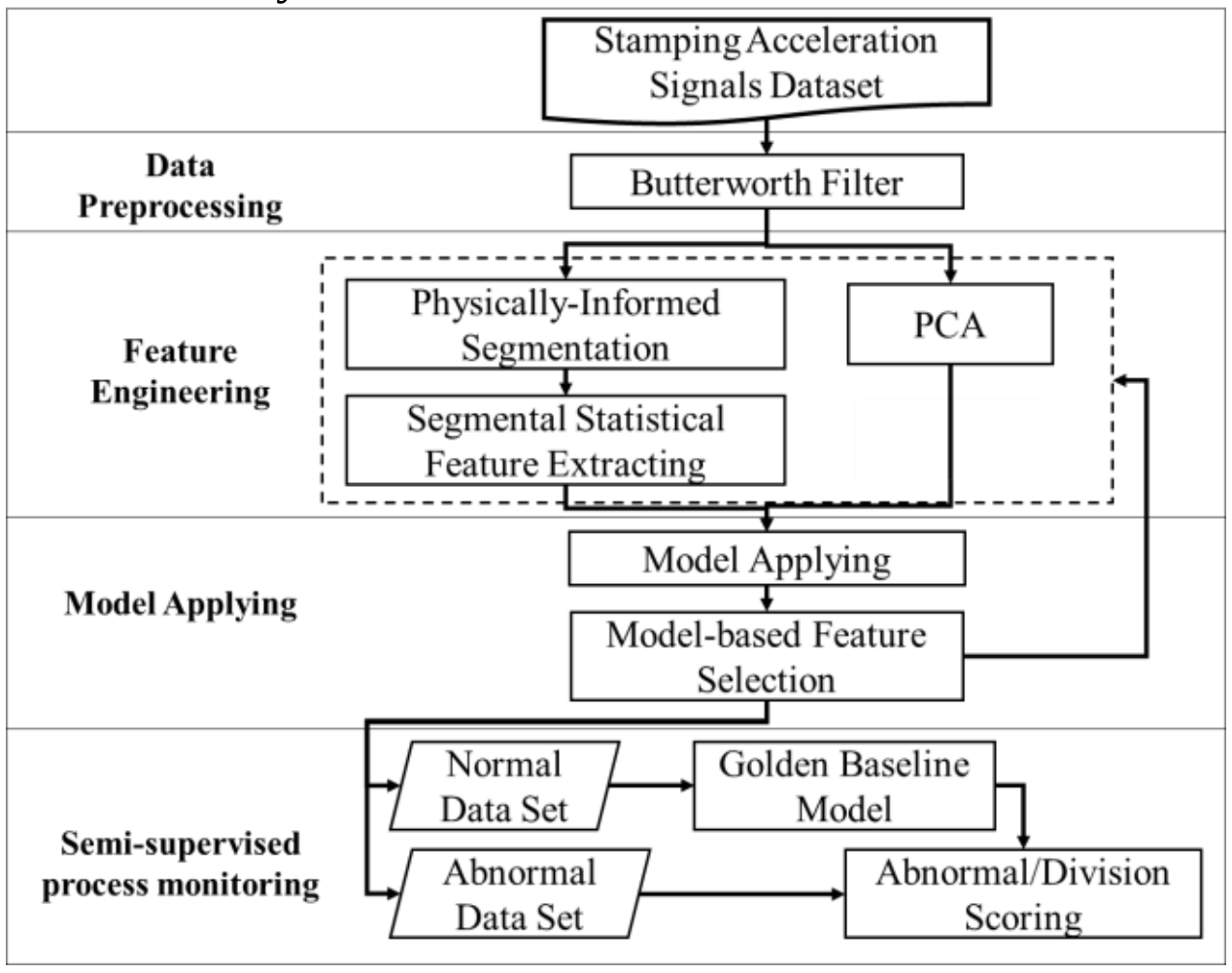

Fig. 1. Flow chart of the proposed stamping processes abnormal monitoring method.

### *2.1. Dataset Information*

The dataset utilized in this study consists of acceleration signals collected from sensors which are affixed to the die set, capturing data at a sampling frequency of 100,000 Hz. The dataset encompasses 1,408 samples, each representing a single cycle of the stamping process and containing approximately 17,000 to 18,000 data points, segmented based on cam signals from 150° to 200°.

Among the total samples, 40 are identified as anomalies, with their corresponding work-pieces marked as defective, establishing a normal to anomaly sample ratio of 30:1. This ratio reflects the inherent characteristics of the stamping process, where, considering a typical high-volume stamping production line, the defect rate generally ranges from 0.5% to 2%, indicating an even more obvious disparity between normal and anomaly samples in a real-world setting.

While the primary objective of this paper is to investigate anomaly detection methods, thus, during the feature extraction phases, subsampling of normal sample set is implemented to mitigate biases in the technical exploration. However, the original ratio of normal to anomaly samples is preserved when building the golden baseline model with semi-supervised learning method.

### *2.2. Data Preprocessing*

While acceleration sensors are preferred for their precision and cost effectiveness, the application in stamping monitoring is challenged by the significant noise in data. The noise, which frequently arises from various sources within the manufacturing scenario, including environment noise in the workshop and operational noise from other machinery, complicates the interpretation of acceleration signals.

Spectral analysis, as depicted in Figure 2, reveals a concentration of signal power within the lower frequency bands, identifying pronounced components between 1800-2500 Hz, 2500-4000 Hz, and 6000-7000 Hz. Given the dynamic nature of the stamping process and typical collision movements, effective frequencies are generally below 7500 Hz[17]. Therefore, to minimize the impact of noise and non-informative frequencies on the signal and to reduce variability in subsequent feature engineering, a filtering approach is necessary.

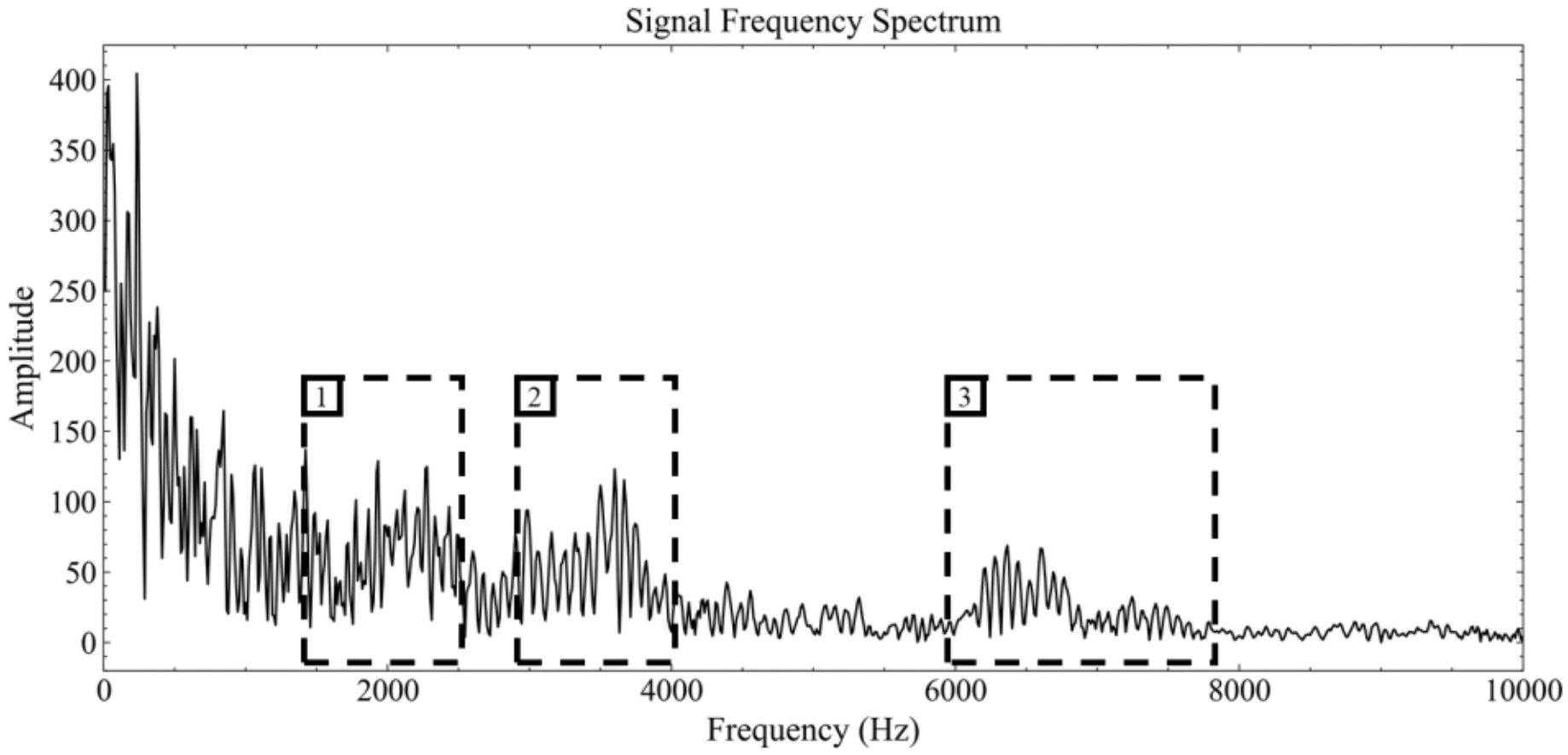

Fig. 2. The frequency spectrum of the raw signal, with the first box ranging from 1800 to 2500 Hz, the second box from 2500 to 4000 Hz, and the third box from 6000 to 7000 Hz.

The Butterworth filter, renowned for its effectiveness in diminishing background and operational noises while retaining fundamental frequency components, is utilized to preserve the essence of the stamping dynamics.

$$|H(\omega)|^2 = \frac{1}{1+\left(\frac{\omega}{\omega_c}\right)^{2n}} \tag{1}$$

where $\omega_c = 2\pi f_c$, the cutoff frequency, $f_c$, marks the threshold in the frequency spectrum where the filter begins to notably attenuate frequencies. The filter's order, $n$, determines the steepness of the filter's roll-off rate past this cutoff point, with a higher order indicating a more rapid decline in signal beyond the cut-off point. Selecting effective parameters is therefore imperative for optimal signal processing, allowing for the reduction of noise without compromising the integrity of the essential frequencies associated with the stamping process.

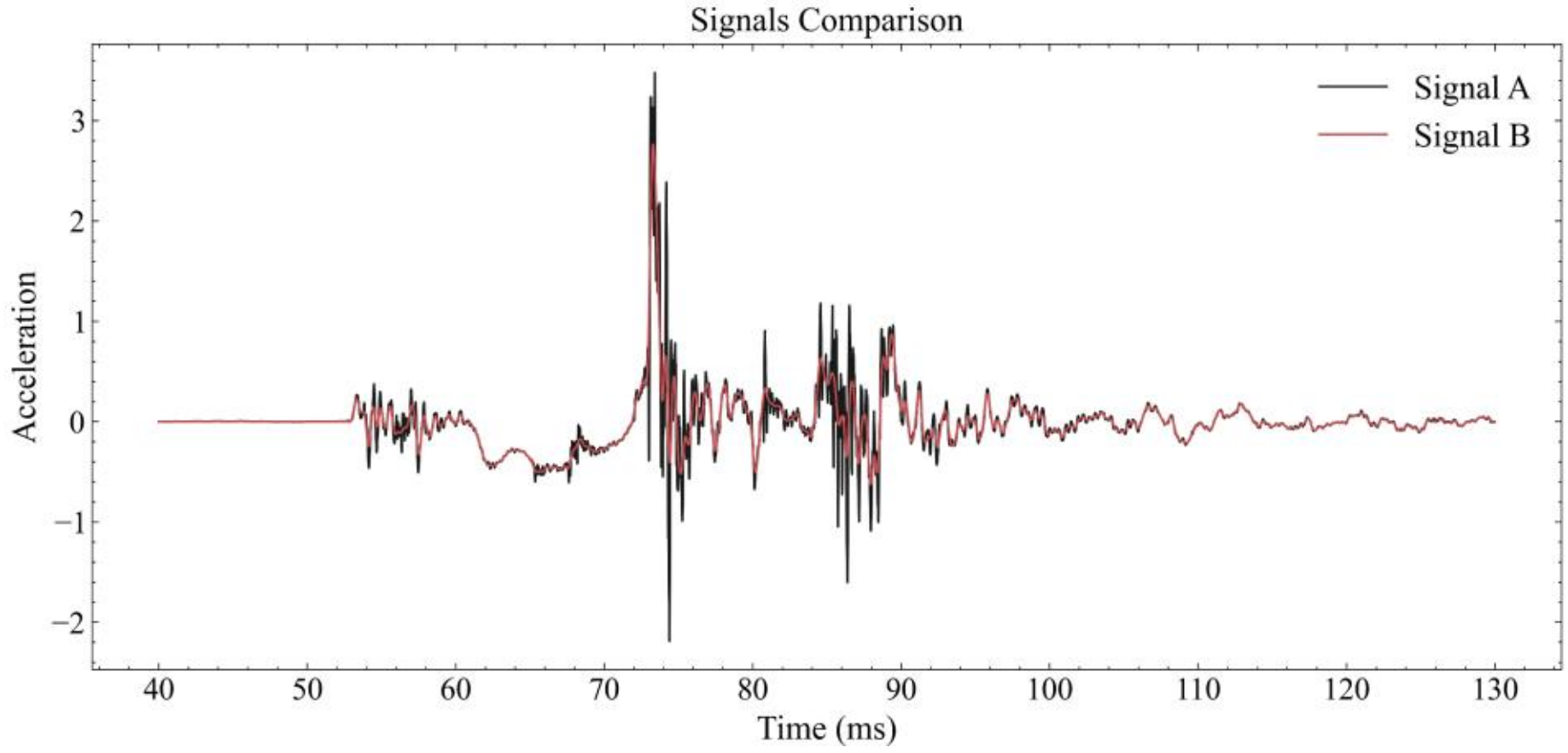

Fig. 3. Comparison of raw signal and filtered signal.

Considering the mechanical characteristics of the stamping process and the frequency analysis, a cutoff frequency of 1800 Hz and an order of 3 are designated for the low-pass filter to effectively suppress extraneous high-frequency noise[18]. This preserves the core signal components crucial to process characterization. Figure 3 illustrates a comparison of a particular signal before and after the application of the low-pass Butterworth filter. It can be observed that post-filtering, the signal exhibits reduced fluctuations, yet substantially retains the majority of waveform information.

Additionally, considering the feature extraction that follows, irregular noise significantly increases the variance between features of samples. As shown in Figure 4, Figure 4a represents samples collected during 10 consecutive normal stamping processes. The signal contains obvious noise, leading to differences between signals that should be similar, thereby complicating the accurate characterization of the stamping process. In contrast, Figure 4b depicts the signal after filtering, where an apparent increase in uniformity is observed. This enhanced consistency of the signals is particularly beneficial for the subsequent stages of feature extraction and analysis.

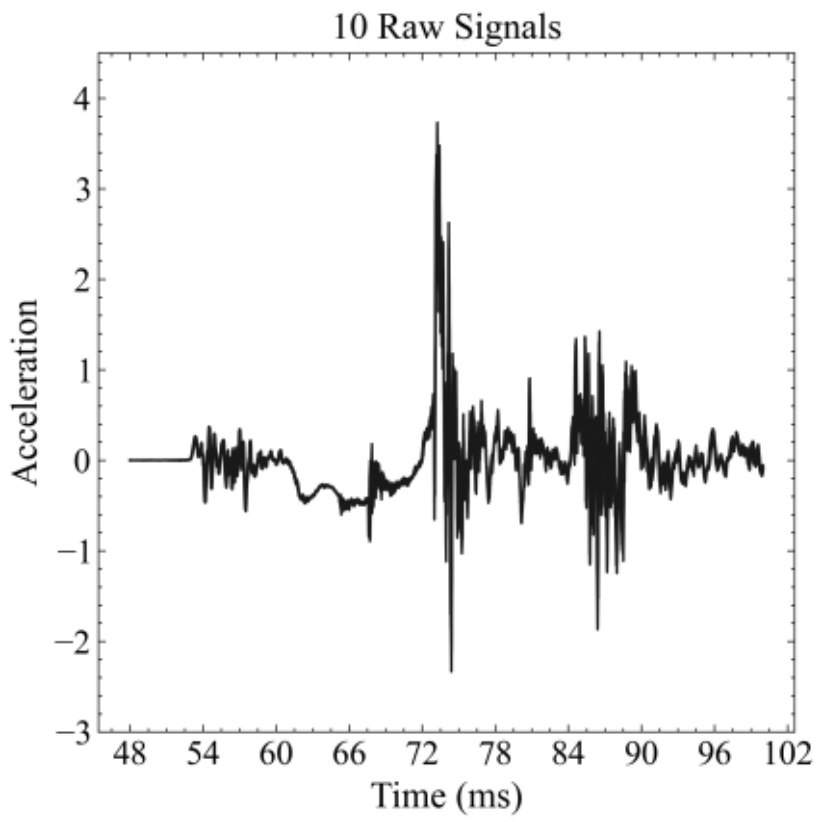

(a)

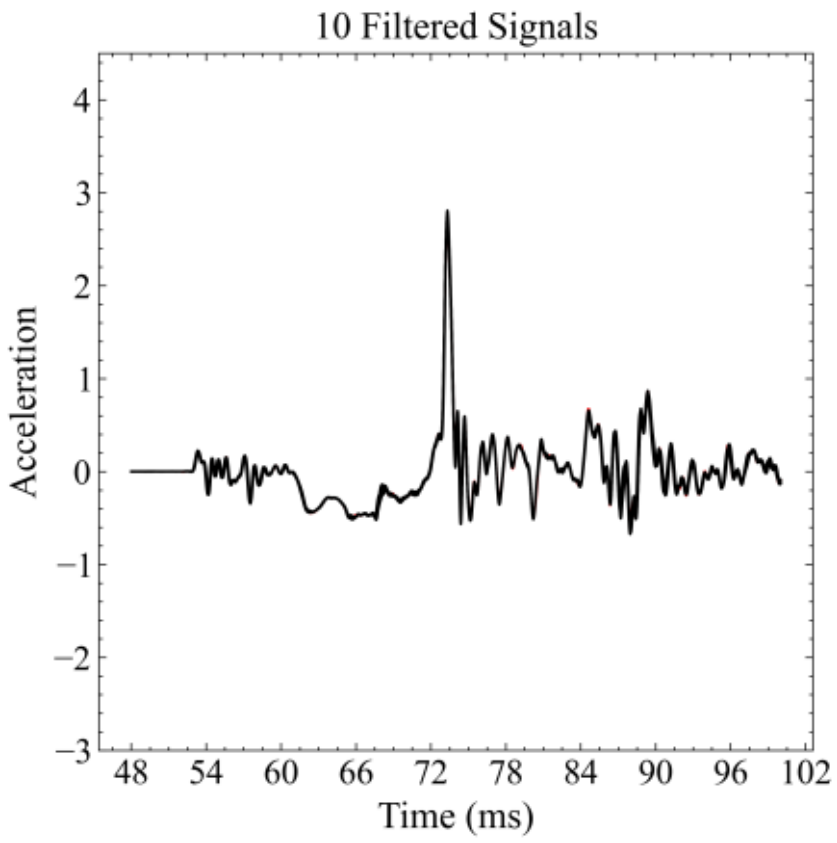

(b)

Fig. 4. Comparison of 10 raw signals(a) and 10 filtered signals(b).

In the preprocessing stage, the deployment of the Butterworth low-pass filter, guided by spectral analysis and empirical insights, which sharpens the raw data by sieving out extraneous noise, substantially enhancing data clarity and reliability. This improvement is pivotal for the data to more accurately reflect the stamping dynamics, crucial for subsequent analysis and decision-making processes.

### *2.3. Feature Engineering*

Given the redundant information from acceleration signals, it is crucial to extract key information while addressing dimensionality challenges. However, due to the complexity of the stamping process, characterized by dynamic interactions and environmental disturbances, relying solely on traditional statistical methods for feature extraction and analysis could not represent the mechanical interactions and dynamics of the stamping process precisely, making it difficult to identify key features. Therefore, analysis based on characteristics specific to the stamping process with combination of physics inform and data-driven approaches is necessary. Furthermore, due to the presence of transient information in the stamping process, time-domain analysis and feature extraction methods tend to be more effective than those based on the frequency-domain. This paper presents a feature extraction framework that integrates physically-informed and data-driven approaches, combining segmental feature extraction and PCA methods.

For the study of stamping process mechanics, researchers have utilized signals such as force, acoustic emissions, and acceleration signals, combined with cam signals or mold displacement curves to describe stamping process and discover knowledge information[19–21]. Some of the studies classify the stamping process into multiple stages based on the motion of the stamping equipment or material states undergoing changes. In this work, the stamping process is divided into seven stages based on six critical points identified in the stamping operation. Stage S1 represents the idling phase before Point A, with minimal signal indicating negligible vibrations. At Point A, when the upper die contacts the work-piece, the process transitions to Stage S2, the die closing phase, characterized by minor mechanical impact and slight vibrations, indicating early interaction. Stage S3, the elastic deformation stage, starts at Point B as the punch presses the work-piece, forcing it into elastic deformation, causing deformation and notable variations. Stage S4, plastic deformation stage, is defined by the stamping force exceeding the material's yield strength at Point C, when the work-piece begins to undergo plastic deformation, leading to rupture and a significant increase in vibration signals, peaking at Point D. Stage S5, the fracture stage, begins with a decrease from Point D as the work-piece fractures and then continues declining at Point E as it releases strain energy. Stage S6 follows, with the punch and die retracting from the work-piece and a corresponding decrease in vibration signals. And with Point F, Stage S6 ends and Stage S7 returns to idling phase.

Figure 5 illustrates the signal progression through distinct stages of the stamping process, providing insights into the mechanical interactions during the process. Seven stages are identified in the analysis of the stamping operation.

The different stages identified in the stamping process are instrumental in comprehending the mechanical interactions during operations. Focused extraction of features from each stage facilitates the isolation of parameters that are vital to the dynamics of stamping. Researchers have extensively explored segmentation and waveform recognition[22–25], particularly using cam signals for refined segmentation in stamping[21]. However, since this paper primarily addresses the effectiveness of features and model construction, the topic of signal segmentation and waveform identification will not be elaborated upon further.

In this context, extracted features, including length, peak to peak (P2P), and energy from Stage S2 to S5, are pivotal in capturing the stamping operation's temporal dynamics. These segmental features form the foundation for developing analytical models aimed at enhancing the stamping process's efficiency.

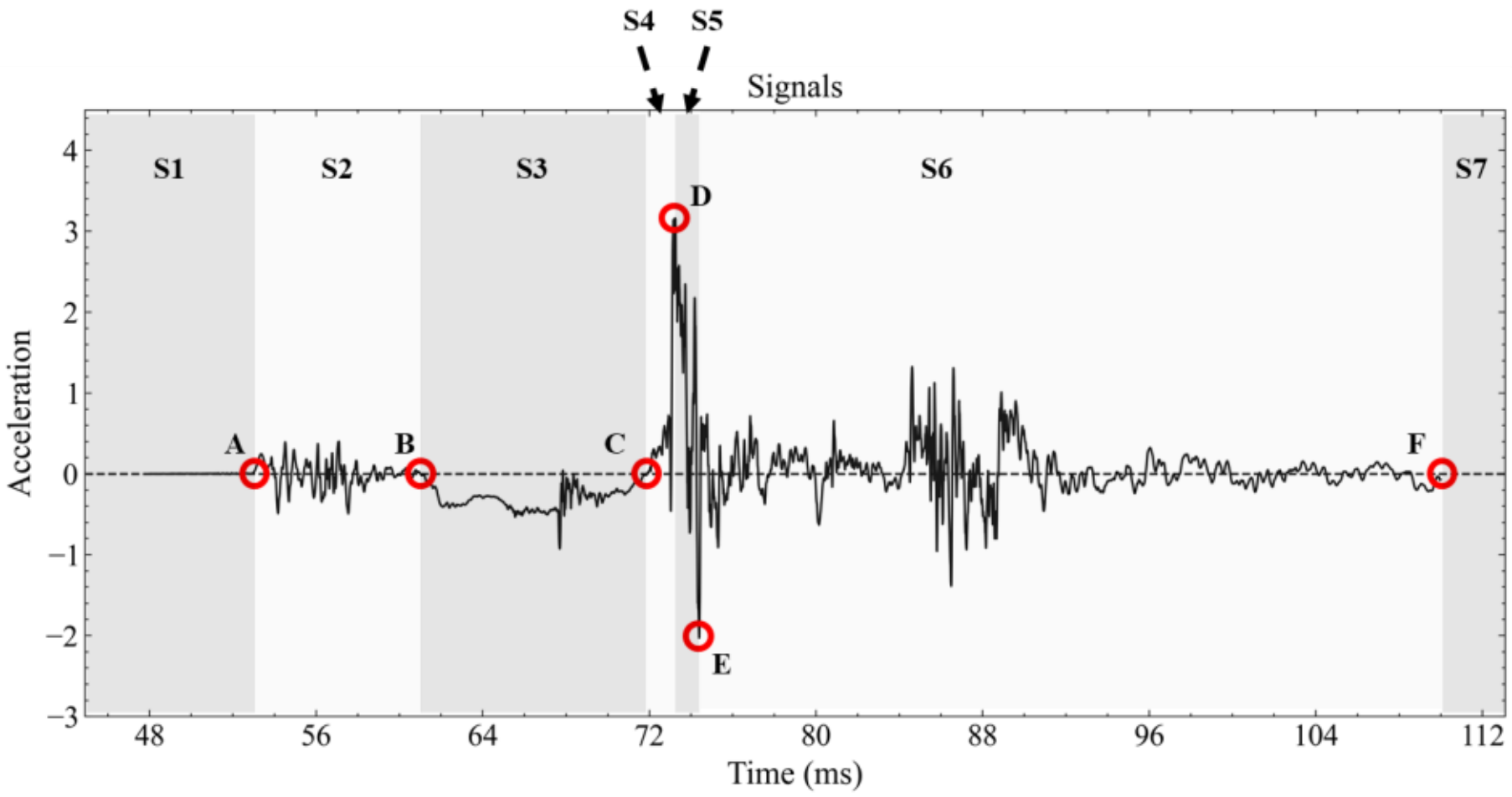

Fig. 5. Signal segmentation based on mechanism information assigns the Stages S1 to S6 from points A to F. Point A represents the critical moment when the upper die first contacts the work-piece, distinguishing between the S1 idling phase and the S2 die closing phase. Point B marks the beginning of the work-piece compression, separating the S2 and S3 elastic deformation stage. Point C signifies the critical point where the stamping force exceeds the material's yield strength, separating the S3 and S4 plastic deformation stage. Point D marks the onset of material fracture, distinguishing the S4 and S5 fracture stage. Point E indicates the end of material fracture, separating the S5 and S6 die retraction stages.

Except for segmental features, PCA, a technique for data dimensionality reduction, is employed to transform the data into a new coordinate system, minimizing dimensions while retaining significant data variability.

It operates by transforming the original data into a set of linearly uncorrelated variables known as principal components, which are ordered by the magnitude of their variance. In constructing the PCA model for stamping process data, the covariance matrix is calculating by:

$$\Sigma = \frac{1}{m-1} Z^T Z \tag{2}$$

where $Z$ is the matrix of standardized data. The eigenvectors $\boldsymbol{v}_j$ and eigenvalues $\lambda_j$ of the covariance matrix are then computed, satisfying the equation:

$$\Sigma \mathbf{v}_j = \lambda_j \mathbf{v}_j \tag{3}$$

Each eigenvalue $\lambda_j$ quantifies the variance along its corresponding eigenvector $\boldsymbol{v}_j$, and the eigenvectors are arranged in descending order of their eigenvalues. The principal component for each observation is then calculated as:

$$C_{ij} = \mathbf{v}_j^T \mathbf{z}_i \tag{4}$$

where $\boldsymbol{C}_{ij}$ is the value of the $i$-th observation on the $j$-th principal component, $\boldsymbol{v}_j^T$ is the transpose of the $j$-th eigenvector, and $z_i$ is the $i$-th standardized original observation.

The determination of which principal components to retain typically involves the balance of data compression and the preservation of variability. Some studies in industrial scenario indicate that the first few components capture the main trends and patterns within the data, which could satisfy most modeling requirements [26]. In this study, the first five principal components are chosen for modeling and analysis of anomaly detection.

*2.4. Model Applying*

As previously noted, to explore a robust anomaly detection technique, the study should utilize a subsampled normal sample set during the model construction and feature selection phases. Thus, the data set is randomly divided into training and testing sets with about 5:2 ratio at first. For the training set, to adjust the ratio of normal to abnormal to a more balanced level, a subset of normal samples is extracted specifically targeting a 10:1 ratio. The composition of the training and testing sets is depicted in Figure 6.

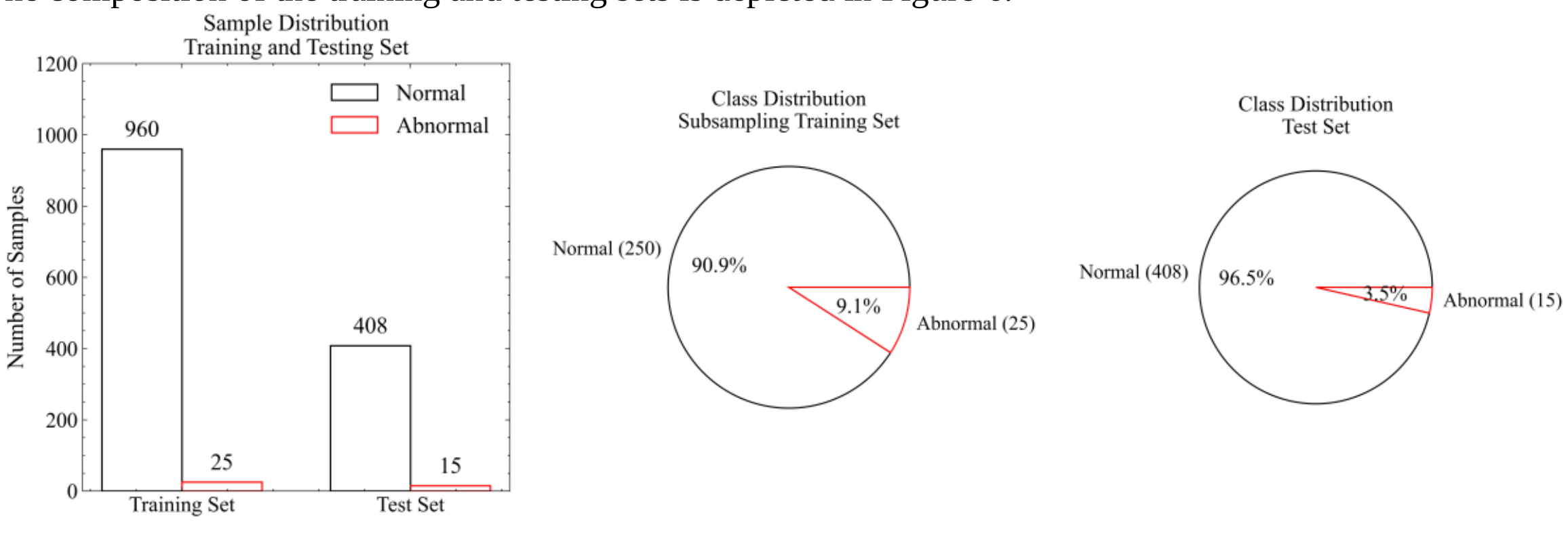

(a) (b) (c)

Fig. 6. Division of training set and test set, with 960 normal 25 abnormal samples in training set and 408 normal 15 abnormal samples in (a); 250 normal 25 abnormal samples in subsampling training set in (b), in which the ratio is about 10:1; 408 normal 15 abnormal samples in testing set in (c).

For the analysis, the feature set is composed of 17 key features, which includes the first five principal components from PCA and twelve segmental features extracted from phases S2, S3, S4, and S5 of the stamping process. Each phase contributes three features, length, P2P and energy, which are critical for capturing the dynamics of the stamping operation. The S6 phase, primarily reflecting vibrations during post-punching stage, is excluded from the feature set as this phase does not involve stamping force and motion which could cause anomalies such as slug. To validate the effectiveness of the selected features, traditional time-domain statistical features such as mean, variance, peak, P2P, energy, and RMS are extracted from signal without segmenting, along with frequency-domain features including spectral density, fundamental frequency, frequency center, bandwidth, and harmonic ratio. These features are used as a control group to construct models as well.

In the development of the stamping monitoring model, several algorithms from statistical methods, ML, and DL are employed to construct models. This approach ensures a comprehensive validation of the preprocessing techniques and feature set effectiveness, irrespective of the type of algorithmic. Specifically, the statistical models include Linear Discriminant Analysis (LDA), Naive Bayes Classification (NBC), and Logistic Regression (LR). The ML category utilizes Support Vector Machine (SVM), Random Forest (RF), and k-Nearest Neighbors (KNN). In the domain of deep learning, a Multilayer Perceptron (MLP) is applied. These models collectively facilitate a robust evaluation of the entire workflow—from data preprocessing to feature engineering and model application—ensuring the technical pipeline's efficacy in stamping process monitoring.

To ensure optimal model performance, grid search is employed for hyper-parameter optimization across all models. This systematic technique scans a predefined space of hyper-parameters, evaluating the effectiveness of each parameter combination to determine the settings that provide the best accuracy. Such meticulous fine-tuning is essential in machine learning to ensure the models are well-adapted to the specific nuances of the dataset, thereby enhancing their robust.

Due to the SVM's sensitivity to the magnitude of input features, a standardized feature set is employed for this model. Feature scaling ensures that the model does not bias towards features with larger scales, allowing the kernel function's computations to reflect the true distances between feature vectors. This step ensures that the range of feature values does not adversely impact the results or the convergence of the algorithm. The same scaling process is also applied to LR, KNN, LDA, NBC, and MLP to enhance training and evaluation.

Grid search is implemented with cross-validation to evaluate various hyper-parameter combinations and optimal parameters are selected based on peak cross-validation accuracy. Then, test dataset is utilized to evaluate the performance of all models which is built with the best hyper-parameters. Results of two feature sets are documented in Table 1.

Table 1. Results of 7 algorithms

| Model | Feature | TP | TN | FP | FN | FPR (%) | FNR (%) |
|---|---|---|---|---|---|---|---|
| LDA | Proposed | 14 | 404 | 4 | 1 | 0.98 | 6.67 |
| | Statistical | 9 | 402 | 6 | 6 | 1.47 | 40.00 |
| NBC | Proposed | 14 | 407 | 1 | 1 | 0.25 | 6.67 |
| | Statistical | 6 | 407 | 1 | 9 | 0.25 | 60.00 |
| LR | Proposed | 13 | 408 | 0 | 2 | 0 | 13.33 |
| | Statistical | 6 | 407 | 1 | 9 | 0.25 | 60.00 |
| KNN | Proposed | 13 | 408 | 0 | 2 | 0 | 13.33 |
| | Statistical | 3 | 407 | 1 | 12 | 0.25 | 80.00 |
| SVM | Proposed | 15 | 408 | 0 | 0 | 0 | 0 |
| | Statistical | 15 | 388 | 20 | 0 | 4.90 | 0 |
| RF | Proposed | 14 | 408 | 0 | 1 | 0 | 6.67 |
| | Statistical | 7 | 408 | 0 | 8 | 0 | 53.33 |
| MLP | Proposed | 15 | 408 | 0 | 0 | 0 | 0 |
| | Statistical | 13 | 408 | 0 | 2 | 0 | 13.33 |

Based on the characters of stamping process, which operates at about 70 strokes per minute, the False Positive Rate (FPR) is particularly critical, as excessive false positives can lead to unwarranted production stops. Equally the False Negative Rate (FNR) also plays an important role, which aims to minimize undetected anomalies to prevent defective outputs. For feature set of segmental and PCA, the performance of the seven models from the statistical, ML, and DL is robust, compared with models with statistical feature set which have quite a few samples of false negative (FN) and false positive (FP). And better performances are recorded of the ML and DL models, with the best two models are SVM and MLP. Particularly, both of them have no FN and FP samples. The results illustrate the effectiveness of the hybrid feature extraction method for stamping process monitoring, with all three types of algorithms are robust.

*2.5. Model-Based Feature Selection*

Among the models employed, the RF algorithm could provide quantification of feature importance which could assist the assessment and selection of features. Additionally, feature selection method based on mutual information is also implemented to enhance the effectiveness of feature selection and evaluation.

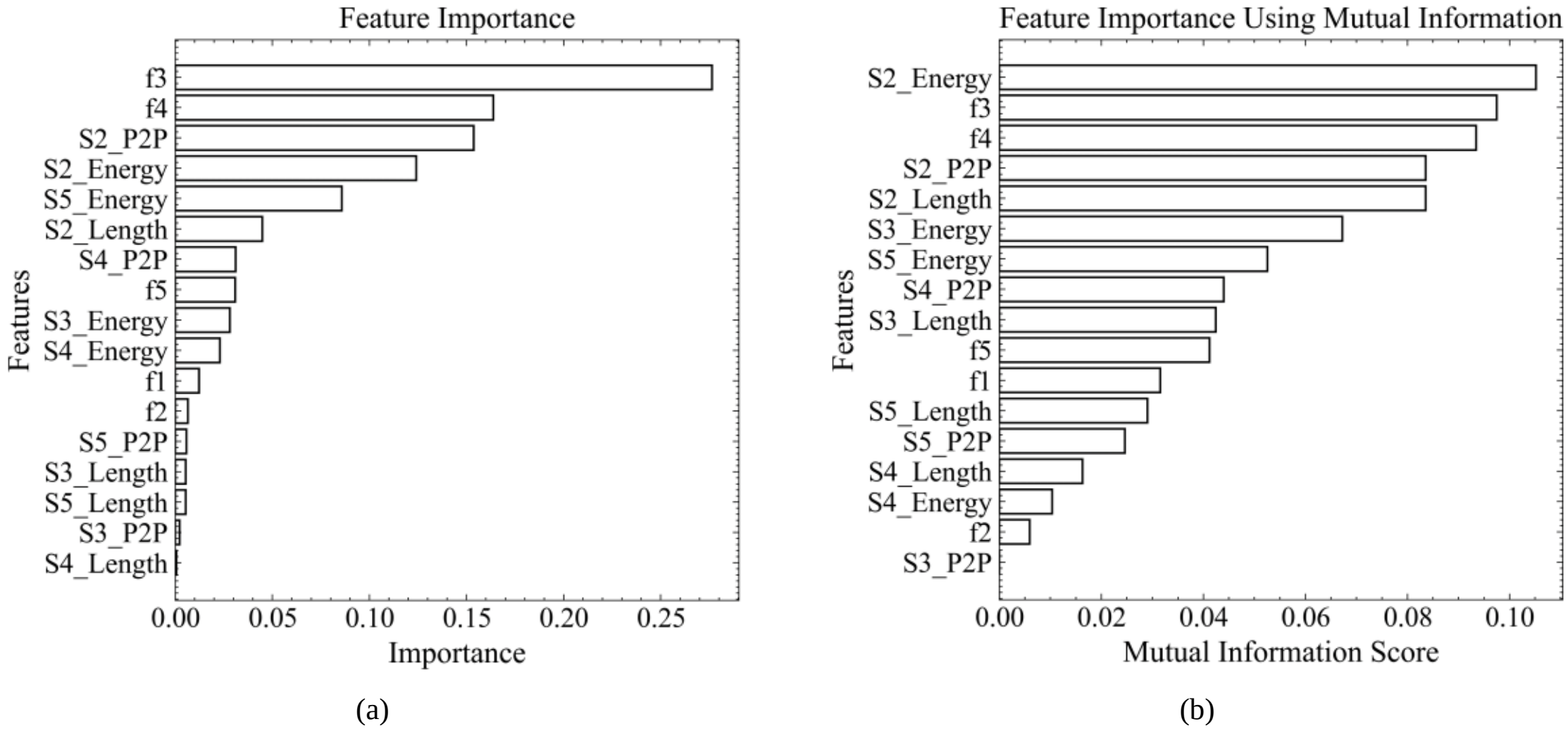

Fig. 7. Feature importance from RF(a) and mutual information(b).

Feature importance analysis, illustrated in Figure 7, emphasizes the significance of specific features. Figure 7a shows the feature importance under optimal RF parameters, while Figure 7b displays results from the mutual information method. Notably, apart from some principal axes identified by PCA, the three statistical features from the S2 phase—length, P2P, and energy—are prominently ranked.

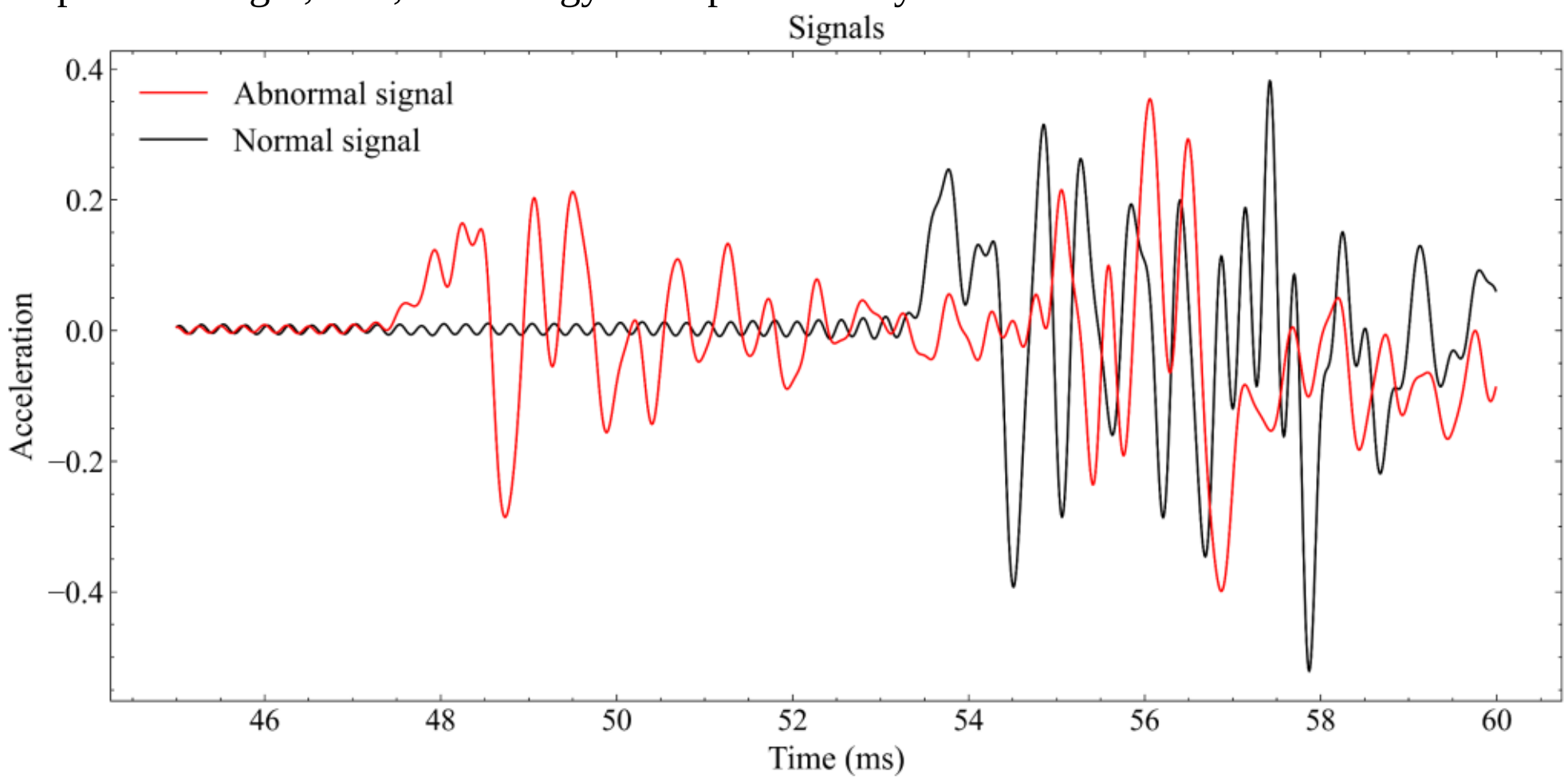

Fig. 8. Comparison of S2 phase between normal and abnormal signals.

Figure 8 provides a targeted analysis of the S2 phase, showing that:

- There is a clear distinction between normal and abnormal samples within the S2 phase, with abnormal samples typically entering S2 phase earlier, and exhibiting lower peak values and longer durations.
- The analysis of the interaction between the upper and lower dies suggests that Point A, the start of the S2 phase, likely represents a critical point between contact and non-contact. And the vibration observed in the S2 phase result from the interaction of impact forces and the kinetic energy of the upper die, which is consistent with physical principles where the presence of anomalies necessitates earlier interactions between the dies than typically expected.

In the analysis of key features within the S2 phase—namely length, P2P, and energy—scatter plot visualizations have revealed insightful correlations. As illustrated in Figure 9, the S2_Length feature notably excels in differentiating between normal and abnormal samples, where a simple demarcation line suffices for clear segregation (Figure 9a and Figure 9b). Moreover, the scatter plots of S2_Energy and S2_P2P distinctly cluster normal and abnormal samples, indicating their potential as reliable indicators for anomaly detection (Figure 9c).

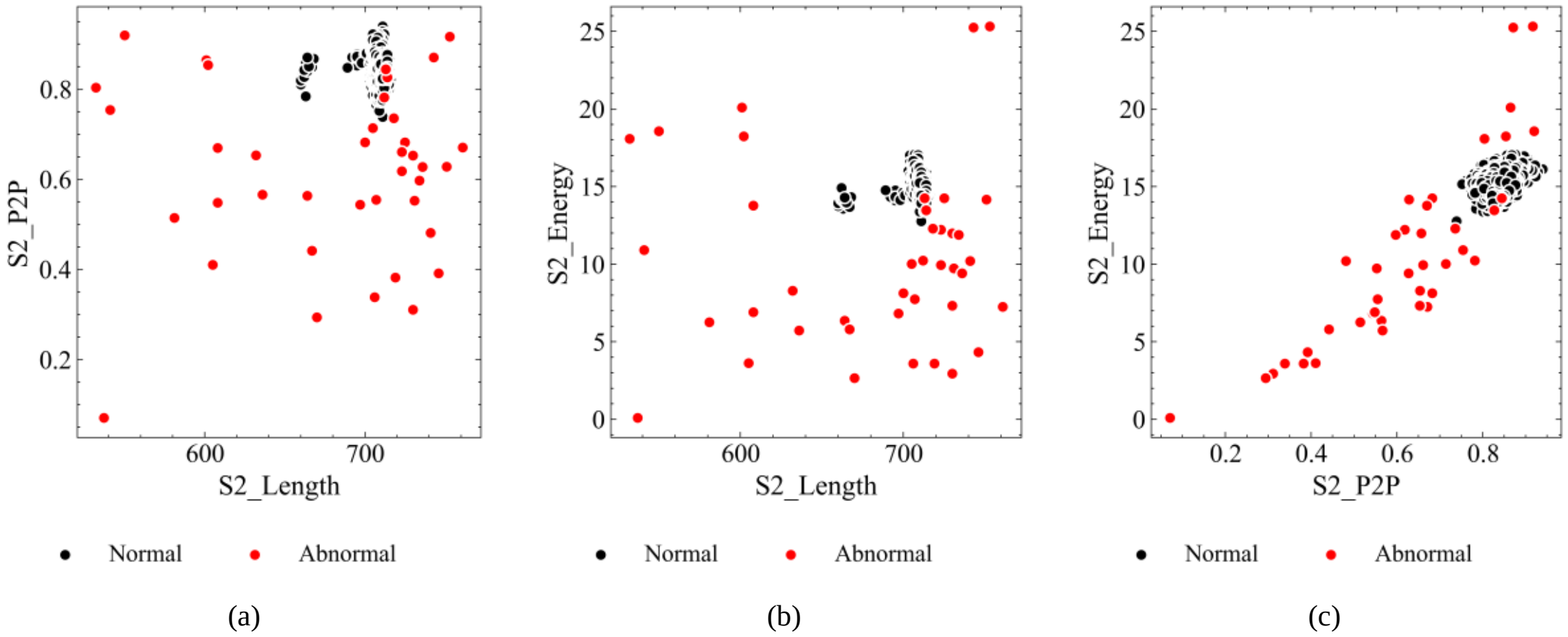

Fig. 9. Comparison of three features based on S2 phase length, P2P, and energy with sample labels.

The analysis of feature importance and their interrelationships has highlighted the significance of the three features from the S2 phase for anomaly detection in the stamping process, underscoring their relevance from both mechanistic and statistical standpoints. Thus, the concept of model-based feature selection is employed to validate this finding.

With the dataset, training, and testing sets maintained constant, feature combinations are adjusted to facilitate the construction of models using diverse sets. Specifically, three distinct feature sets are utilized: the S2 feature set, which includes only the three features from the S2 phase; the optimal feature set, which consists of the top features identified by both the Random Forest and mutual information analyses including S2_Length, S2_P2P, S2_Energy, S3_Energy, S4_P2P, S5_Energy, f3, and f4; and the statistical feature set. Models are constructed using seven algorithms from statistics, machine learning, and deep learning categories. After optimizing parameters through grid search, results are compiled in Figure 10.

Upon comparing the outcomes from models utilizing all four feature sets, it is evident that models incorporating all features, optimal features, and S2 features generally exhibited superior performance. This demonstrates that the three features of the S2 phase exhibit some key information, which could notably characterize the stamping process, especially in detecting anomalies like slugs. However, given the limited size of the current dataset, further validation of these conclusions is necessary. Furthermore, based on these findings, continued exploration and feature extraction within the S2 phase are warranted to advance the quantification of anomalies in the stamping process.

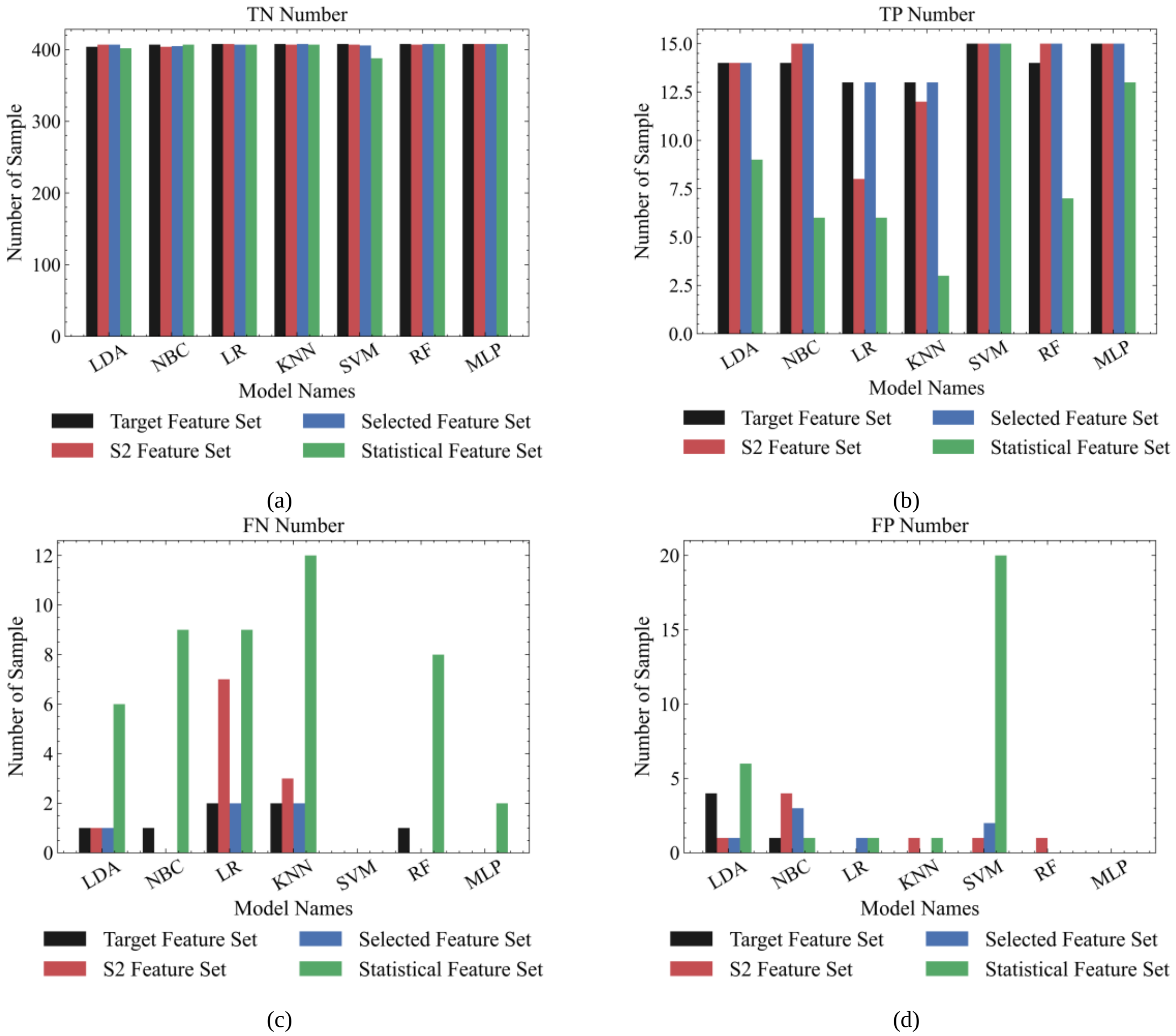

Fig. 10. Results of 7 algorithms with different feature sets.

## 3. Application to Stamping Process Monitoring

### *3.1. Unbalanced Normal Abnormal DataSet*

In the context of stamping process, the dataset exhibits a significant imbalance between normal and abnormal samples, with a current ratio of 30:1, which is even more obvious in real-world scenarios. Currently, yield requirements for most factories are very stringent, especially in high-speed, large-scale stamping processes where the yield is generally expected to be controlled within 1%, which leads to an extremely exaggerated ratio of positive to negative samples. Additionally, in high-speed stamping processes where strokes per minute range from several dozen to thousands, and due to the closed nature of the production lines, it is challenging to obtain ideal labels and even anomaly samples during the early stages of monitoring system development. Considering these two factors, the imbalance of positive and negative samples presents a significant challenge that must be addressed for monitoring algorithms and systems during the initial phases of deploying.

Industry practitioners and academic researchers tackle this imbalance by applying a variety of techniques designed to enhance model training and improve detection accuracy. Some scholars focus on data-level strategies to balance the distribution between classes, which include random under-sampling, over-sampling and hybrid techniques that combine both approaches to ensure a balanced dataset without biasing towards the majority class[27]. However, this method does not effectively resolve the issue, particularly as it fails to make effective use of the large number of normal samples.

*3.2. Golden Baseline Model Building*

In this context, given the challenges of obtaining anomaly samples and based on aforementioned research on stamping data and features, a one-class algorithm is employed for anomaly detection under conditions of data imbalance. This approach utilizes the abundant normal data, constructing a one-class boundary based on the similarities to represent the normal data sample space. For new samples, the distance to this boundary is calculated, and an anomaly is identified if this distance exceeds a predefined threshold. By this approach, normal samples are effectively distinguished between normal samples and the few outliers. For the given set of all normal samples, a mathematical decision boundary is utilized to characterize and encompass the normal data. The decision function is represented as follows:

$$f(x) = \langle \boldsymbol{w}, \phi(\boldsymbol{x}) \rangle + b \tag{5}$$

where $\phi(\boldsymbol{x})$ represents the feature mapping function, $\boldsymbol{w}$ is the weight vector, and $b$ is the bias term. Subsequently, to encompass normal data samples, the samples closest to the boundary within the normal data are identified, and the distance between these samples and the boundary should be maximized to enhance the model's generalization performance. Therefore, this is achieved through the following optimization problem:

$$\min_{w,\xi_i,\rho} \frac{1}{2} \| \boldsymbol{w} \|^2 + \frac{1}{\nu n} \sum_{i=1}^{n} \xi_i - \rho \tag{6}$$

subject to:

$$\left\langle \boldsymbol{w}, \phi\left(\boldsymbol{x}_i\right) \right\rangle \geq \rho - \xi_i, \quad \xi_i \geq 0, \quad i = 1, \ldots, n \tag{7}$$

where $\xi_i$ is the slack variables, allowing for some flexibility in the separation to handle noise and outliers, $\rho$ is the decision function threshold, defining the distance from the origin to the decision boundary, $\nu$ is a parameter that controls the trade-off between maximizing the distance of the decision boundary from the origin and the number of outliers allowed on the wrong side of the decision boundary.

To solve this optimization problem, the Lagrange multipliers method is employed to transform the problem into its dual form. By utilizing kernel functions $K(\mathbf{x}_i, \mathbf{x})$, the definition of the distance from the samples to the decision boundary can be determined as follows:

$$d = \sum_{i=1}^{n} \alpha_i K\left(\boldsymbol{x}_i, \boldsymbol{x}\right) \tag{8}$$

where $\alpha_i$ is Lagrange multipliers. When confronted with new data, with specific scores for each sample, an appropriate threshold $\rho$ is defined. If the distance from the decision boundary exceeds the threshold, the sample is classified as an anomaly, the sample is classified as normal. Conversely, if the distance is below the threshold, the sample is classified as an anomaly.

For the study, the dataset comprising 1,408 normal samples and 40 anomalies is divided into three subsets: a training set with 820 normal samples; a validation set including 294 normal and 20 abnormal samples; and a test set with 294 normal and 20 abnormal samples. The training set is used to train the model, while the validation set aided in fine-tuning the parameters to achieve the best performance. The efficacy of the model, with the optimal parameters found on the validation set, is then evaluated on the test set, with the results detailed in Figure 11. The results indicate that the proposed one class model effectively identified most anomalies in the test set with only one FP samples and one FN sample. This suggests that under the conditions of the selected features and the dataset, the one class model performs efficiently.

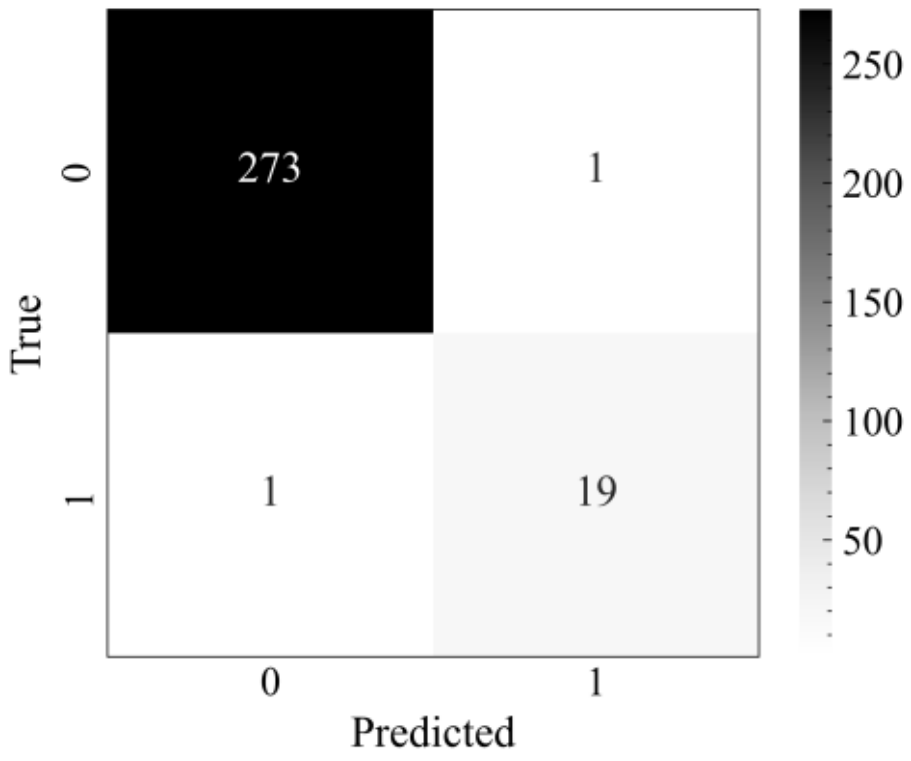

Fig. 11. Results of one class model with optimal feature sets.

### *3.3. Discussion*

In real-world stamping scenario, considering some slight defects may not be defined as anomalies, it is necessary for a monitoring model that allows operational personnel to adjust the anomaly threshold according to specific standards. Furthermore, adjustments to the model may also be essential in the face of changes in materials, process parameters, and drifts of equipment conditions[28,29]. Moreover, quantification of anomalies is also required in this adjustment process.

For the proposed one-class model, the distance, which indicates the proximity of the sample to the decision boundary, can be used as a specific score to assess the quantification of anomaly. Additionally, the threshold $\rho$ , used to determine whether a sample is anomalous, can serve as an adjustable parameter for on-site operators to fine-tune the detection sensitivity based on real-time production dynamics. To better represent the metric of samples and support decision-making, a transformation is implemented as defined below, with normal samples approaching 0 and anomalies approaching 1. This transformation is constructed using the training set and the validation set containing anomalies, enabling new samples to be fitted into this metric. The presentation of the test set is illustrated in Figure 12.

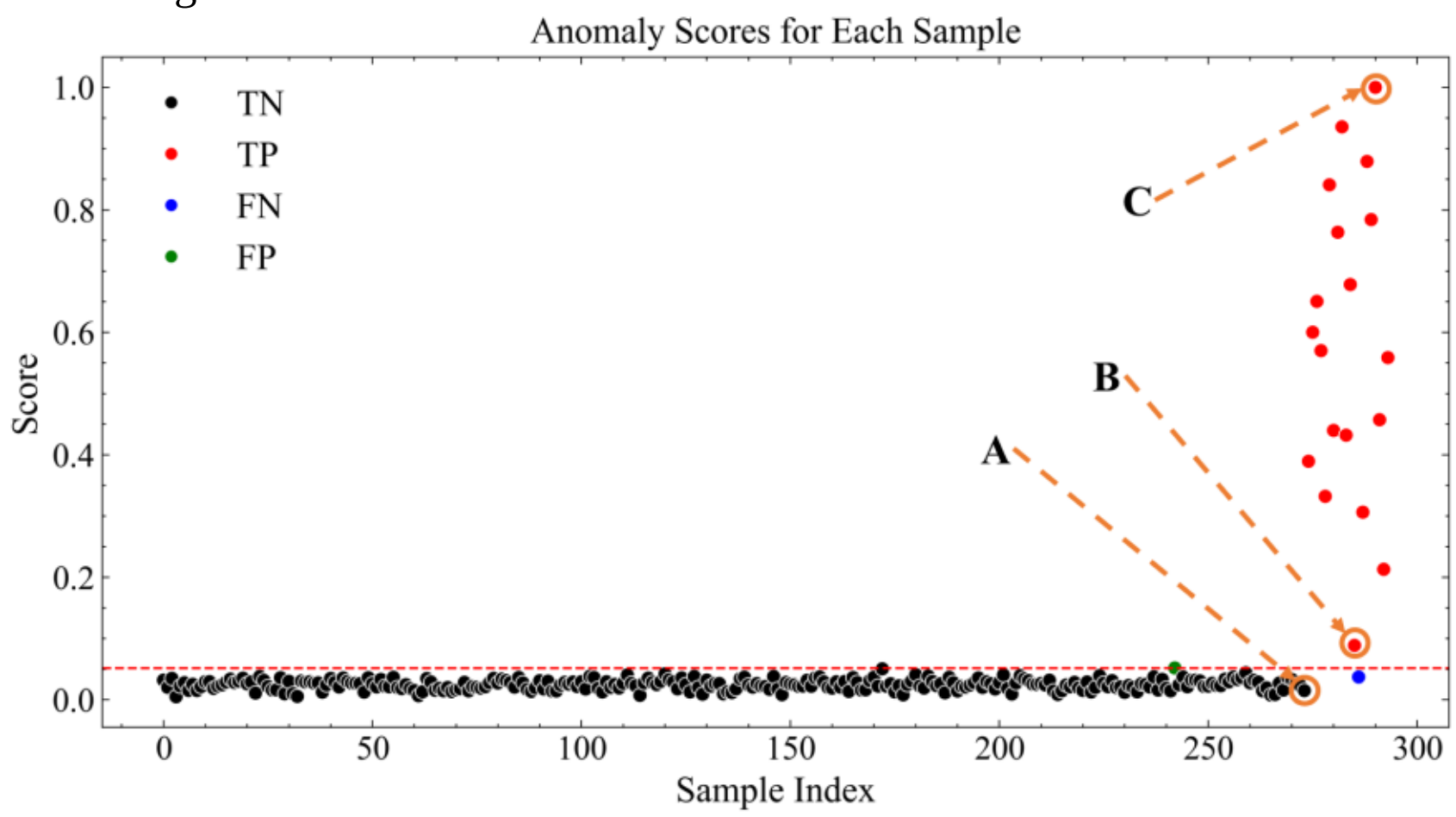

Fig. 12. Anomaly metric for each sample.

As shown in Figure 12, black dots representing TN and red dots representing TP are clearly separated by a red dashed line, while blue and green dots represent FN and FP, respectively. In the diagram, the red dots, due to varying quantification of anomaly, are positioned at different distances from the red decision line along the y-axis. The signals for samples A, B, and C, as shown in Figure 13, have been filtered and time-aligned to remove nonlinear elements and other interferences, facilitating an intuitive observation in the graph. From Figure 13b and 13c, it is evident that compared to the normal signal A, signal B is closer in amplitude to the normal signal, while the effective starting point of signal C occurs earlier than that of signal B. This clearly demonstrates that the anomaly level of sample C is higher than that of sample B.

Based on these two figures, an application can be constructed, allowing personnel to adjust this threshold to fine-tune the model's decisions to meet the specific conditions and quality requirements of the production scenario during operation.

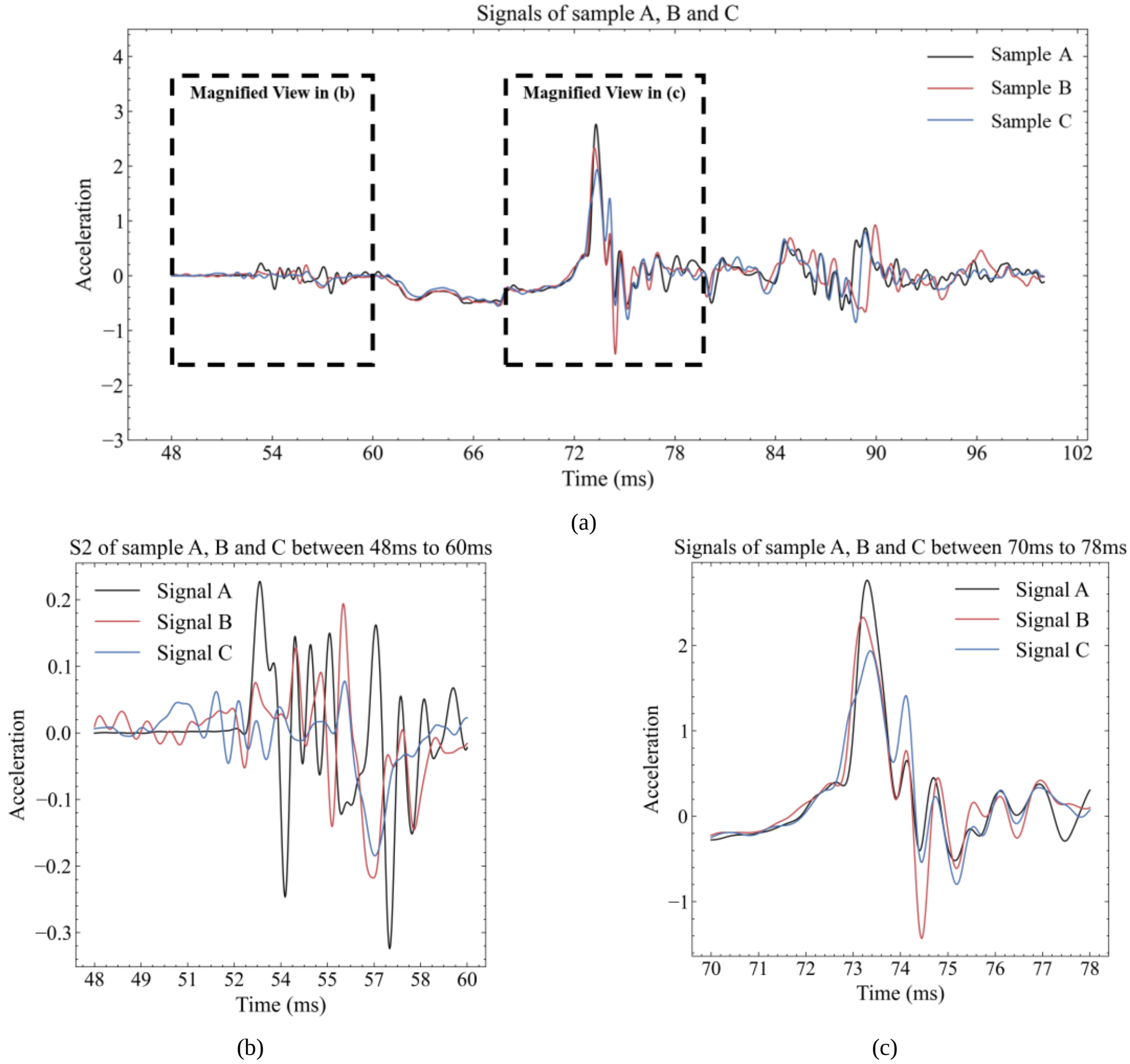

Fig. 13. Signals of sample A, B and C, with the entire effective signal segment (a), signals between 48ms to 60ms (b), and signals between 70ms to 78ms (c).

## 4. Conclusions

This research introduced an enhanced semi-supervised in-process monitoring methodology with a hybrid feature extraction approach, particularly applied in stamping processes for real-time anomaly detection. The proposed framework optimally leverages physics information to facilitate the key feature extraction and selection. Addressing severely imbalanced sample data distribution issues in real-world scenarios, a one-class monitoring model was developed to capture and quantify process anomalies effectively.

The proposed methodology has the potential to make huge positive impact in advanced manufacturing processes where the in-process anomaly detection requirement is demanding. In future work, more data-driven methods and applications across various scenarios will be investigated to enhance its usability and generalizability. Concurrently, to ensure the model's long-term performance, in-depth research on online model updating mechanism will be conducted.